\documentclass[11pt,a4paper]{article}
\usepackage[utf8x]{inputenc}
\usepackage[T1]{fontenc}
\usepackage{amsmath}
\usepackage{amssymb}
\usepackage{gbt7714}
\usepackage{wrapfig}
\usepackage{subcaption} 
\usepackage{framed}
\usepackage{booktabs}
\usepackage{picins}

\usepackage[pdftex]{graphicx} 
\usepackage[pdftex,linkcolor=black,pdfborder={0 0 0}]{hyperref} 
\usepackage{calc} 
\usepackage{enumitem} 

\usepackage[a4paper, lmargin=0.09666\paperwidth, rmargin=0.09666\paperwidth, tmargin=0.1111\paperheight, bmargin=0.1111\paperheight]{geometry} 

\usepackage[all]{nowidow} 
\usepackage[protrusion=true,expansion=true]{microtype} 

\title{Learning to Solve Combinatorial Optimization under Positive Linear Constraints via Non-Autoregressive Neural Networks}

\author{Runzhong WANG, Yang LI, Junchi YAN\footnote{This is the English version of \href{https://www.sciengine.com/SSI/doi/10.1360/SSI-2023-0269;JSESSIONID=c219ff88-b005-463b-9968-7c72c6511e55}{\underline{the paper published on \emph{Scientia Sinica Informationis}}}. Please cite this paper as ``Wang et al, Learning to Solve Combinatorial Optimization under Positive Linear Constraints via Non-Autoregressive Neural Networks. \emph{Scientia Sinica Informationis} (2024)''. Submitted 2023-09-17; Accepted 2024-06-17; English version 2024-09-05. Corresponding to: Junchi YAN (yanjunchi@sjtu.edu.cn)}, Xiaokang YANG\\Shanghai Jiao Tong University}
\date{}

\hypersetup{ 	
pdfsubject = {},
pdftitle = {},
pdfauthor = {}
}

\begin{document} 
\maketitle

\abstract{Combinatorial optimization (CO) is the fundamental problem at the intersection of computer science, applied mathematics, etc. The inherent hardness in CO problems brings up challenge for solving CO exactly, making deep-neural-network-based solvers a research frontier. In this paper, we design a family of non-autoregressive neural networks to solve CO problems under positive linear constraints with the following merits. First, the positive linear constraint covers a wide range of CO problems, indicating that our approach breaks the generality bottleneck of existing non-autoregressive networks. Second, compared to existing autoregressive neural network solvers, our non-autoregressive networks have the advantages of higher efficiency and preserving permutation invariance. Third, our offline unsupervised learning has lower demand on high-quality labels, getting rid of the demand of optimal labels in supervised learning. Fourth, our online differentiable search method significantly improves the generalizability of our neural network solver to unseen problems. We validate the effectiveness of this framework in solving representative CO problems including facility location, max-set covering, and traveling salesman problem. Our non-autoregressive neural solvers are competitive to and can be even superior to state-of-the-art solvers such as SCIP and Gurobi, especially when both efficiency and efficacy are considered. Code is available at \url{https://github.com/Thinklab-SJTU/NAR-CO-Solver}}

\section{Introduction}

Combinatorial optimization (CO) refers to a class of non-convex optimization problems with discrete decision spaces, named for its decision space often "exploding combinatorially" with increasing parameters. As a shared problem in computer science, applied mathematics, and management science, CO has a long research history, e.g., Euler~\cite{euler1741solutio} in the 18th century initiated the study of graph theory, and the pioneering big names such as Birkhoff~\cite{birkhoff1946tres}, von Neumann~\cite{von1953certain}, and Dantzig~\cite{dantzig1951application} systematically studied CO in the 1940s-1950s. Most researchers agree that apart from a few "simple" problems, most CO problems are NP-hard and thus cannot be solved exactly in polynomial time. Nonetheless, with the growth of computational power and machine learning advancements, improving the performance boundaries of current solvers remains an active research direction.

\begin{figure}[!t]
\centering
\includegraphics[width=\textwidth,page=1]{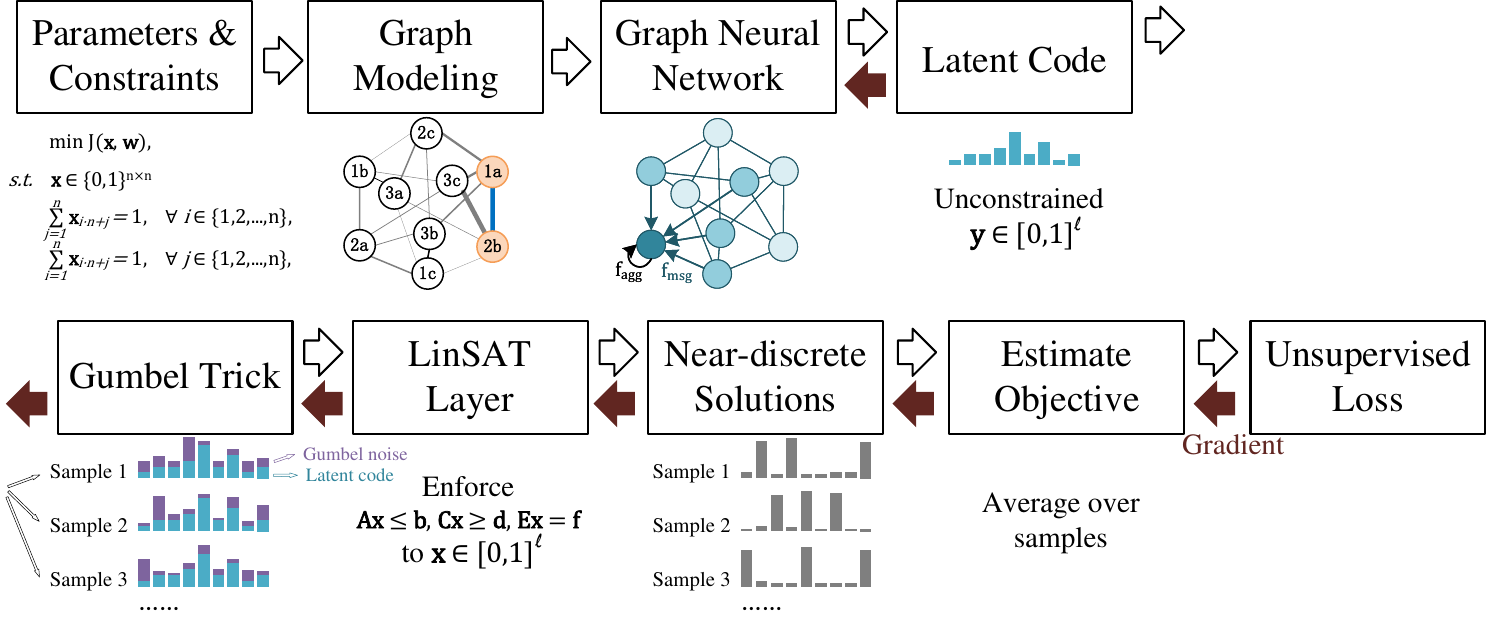}
\caption{Non-autoregressive neural network paradigm proposed to solve combinatorial optimization problems (in training).}
\label{fig:overview_train}
\end{figure}

In particular, Turing Award laureate Prof.\ Yoshua Bengio noted that fitting and solving CO problems under a specific distribution is often easier than developing a general-purpose solver~\cite{BengioEJOR21}. Compared to traditional solvers that typically run on CPUs~\cite{wangfudongPhDThesis,zhouyalan}, neural network algorithms running on GPUs can fully exploit the power of parallelism. Moreover, due to their ability to fit data distributions, neural networks hold significant potential as CO solvers. However, how to design a neural CO solver with the expected performance improvement remains an open problem~\cite{GuoCombBook,GuoSurvey2019}, with the primary challenge being ensuring that the network output satisfies specific discrete constraints.

A common solution is to build autoregressive neural networks, which output solutions step-by-step, applying rules at each step to ensure the final solution lies in the feasible domain~\cite{VinyalsNIPS15,KhalilNIPS17}. While autoregressive networks have gained wide application in CO due to their generality, they face challenges such as error accumulation, large action spaces, sparse reward signals, and the difficulty in modeling the permutation invariance of the problem space.

This paper introduces a non-autoregressive neural network architecture as a CO solver. The non-autoregressive network outputs all decision variables in one forward pass, avoiding the issues of autoregressive models. Additionally, non-autoregressive networks are a mature architecture with proven efficiency and accuracy in fields such as computer vision~\cite{HeResNet,RedmonCVPR16yolo,zhang2020refinedet}. The latest theoretical research also shows that neural networks (especially graph neural networks that break symmetry with random features) can solve CO problems, such as mixed-integer programming~\cite{ChenICLR2023MIP}. To tackle the technical challenges (i.e., neural network outputs are typically unconstrained, while CO problems require the network to output constrained solutions), this paper introduces the Linear Satisfiability Network (LinSATNet)~\cite{WangICML23}, which projects network outputs into the feasible domain defined by positive linear constraints:

\begin{equation}
    \mathbf{A}\mathbf{x} \leq \mathbf{b}, \mathbf{C}\mathbf{x} \geq \mathbf{d}, \mathbf{E}\mathbf{x} = \mathbf{f}, \qquad \text{where} \ \mathbf{A},\mathbf{b},\mathbf{C},\mathbf{d},\mathbf{E},\mathbf{f}\geq 0, \ \mathbf{x}\in[0, 1]^l.
    \label{eq:positive_linear_constraints}
\end{equation}
This covers many common CO problems, such as the knapsack problem, where constraints $\mathbf{w}^\top \mathbf{x} \leq m$ can be handled as "positive linear constraints."

\begin{figure}[!t]
\centering
\includegraphics[width=1\textwidth,page=2]{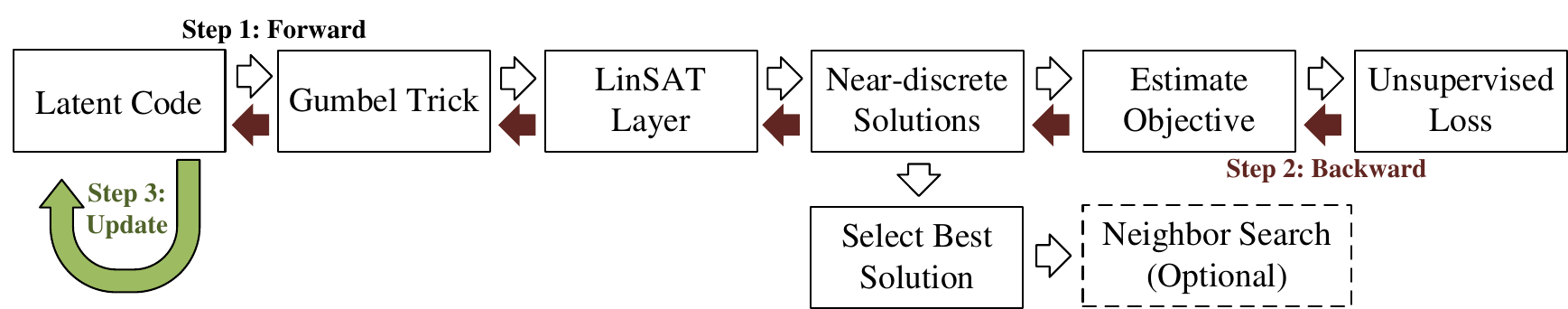}
\caption{Gradient-based optimization steps in non-autoregressive neural network combinatorial optimization solvers (inference).}
\label{fig:overview_test}
\end{figure}

As shown in Figure~\ref{fig:overview_train}, we propose a non-autoregressive neural network paradigm for efficiently solving CO problems. Starting from problem parameters and constraints, the problem is transformed into a graph that can be processed by a graph neural network (GNN). The GNN outputs latent variables (latent code) with the same dimension as the decision variables, which are projected into feasible solutions through the LinSAT layer. With the help of Gumbel reparameterization~\cite{JiangICLR17}, we ensure the network outputs are approximately discrete and feasible. In the inference stage, as shown in Figure~\ref{fig:overview_test}, we further optimize the initial network output through gradient-based search.

The technical advantages of our method are:
\begin{itemize}
    \item \textbf{Generality}: By leveraging LinSAT, our approach can handle a wide range of CO problems with positive linear constraints.
    \item \textbf{Efficiency}: Non-autoregressive networks only require $\mathcal{O}(1)$ forward passes, while autoregressive networks typically require $\mathcal{O}(l)$ passes.
    \item \textbf{Unsupervised Learning}: Our framework supports offline unsupervised learning, reducing the need for labeled data.
    \item \textbf{Differentiable Search}: In the inference stage, gradient-based search can further improve solution quality.
\end{itemize}

We validate the non-autoregressive network's performance on problems such as facility location, max-set covering, and the traveling salesman problem. The proposed solver matches or exceeds traditional solvers like SCIP and Gurobi in terms of efficiency and solution quality.

\section{Related Work}

We follow the survey~\cite{peng2021graph} to categorize neural CO solvers into autoregressive and non-autoregressive.

\textbf{Autoregressive Neural Solvers for Combinatorial Optimization.} Autoregressive neural networks are widely used in sequence learning, where the output at time $(t+1)$ depends on the output at time $t$. In the context of CO, autoregressive neural solvers gradually construct a complete solution step by step. Thus, for problems where decision variables have a dimension of $l$, autoregressive neural networks need $\mathcal{O}(l)$ steps. The advantage of autoregressive networks lies in their ability to restrict the action space at each step, eliminating invalid solutions. This paradigm was first applied to CO in~\cite{VinyalsNIPS15}. Later, Khalil et al.~\cite{KhalilNIPS17} showed that this multi-step decision process could naturally be modeled as a Markov decision process, allowing reinforcement learning algorithms to train the network~\cite{MnihDQN2013}. Due to their flexibility, the "autoregressive network + reinforcement learning" paradigm has become a mainstream research direction, with applications in scheduling~\cite{ZhangNIPS20}, task assignment~\cite{MaoSIGCOMM19}, quadratic assignment~\cite{LiuArxiv20}, bin packing~\cite{DuanAAMAS19}, and more. However, autoregressive networks are prone to accumulating errors over long decision sequences and struggle to model permutation invariance, which may limit their performance and generalization.

\textbf{Non-Autoregressive Neural Solvers for Combinatorial Optimization.} Many CO problems are offline and unrelated to sequential processes. Therefore, non-autoregressive networks, which are widely adopted in other machine learning tasks, seem to be a more natural choice for CO. As early as the 1980s, Hopfield networks were proposed to solve the traveling salesman problem (TSP)~\cite{hopfield1985neural}, using neural networks and gradient descent for direct search, demonstrating the advantages of search in continuous space. More recently, Karalias et al.~\cite{KaraliasNIPS20} proposed a more efficient non-autoregressive neural network, where constraints are handled as penalty terms during training. However, this approach struggles to strictly enforce constraints, and learned constraint information often fails to generalize to new problems~\cite{WangICLR23}. Our prior study~\cite{WangICLR23} further explored different network designs and derived theoretical upper bounds on constraint violations, experimentally verifying that higher violation rates lead to worse neural solver performance. 

This paper demonstrates that the gradient-based search capabilities of non-autoregressive networks provide an advantage, allowing the solver to search for better solutions during inference. The work by Qiu et al.~\cite{qiu2022dimes} emphasizes the importance of gradient search for CO problems. Though they used the higher-variance REINFORCE technique~\cite{williams1992REINFORCE} for constraint handling, their approach differs from ours, which combines Gumbel reparameterization and the LinSAT layer.

\section{Non-Autoregressive Neural Solvers for Combinatorial Optimization}

\subsection{Problem Formulation}
Unless otherwise stated, lowercase bold letters mean vectors, and uppercase bold letters mean matrices. The non-autoregressive network framework proposed in this paper solves binary combinatorial optimization problems with the following form of positive linear constraints:
\begin{subequations}
\label{eq:formulation}
\begin{align}
    &\min_\mathbf{x} \ J(\mathbf{x}, \mathbf{w}),\\
    s.t. \quad & \mathbf{A}\mathbf{x} \leq \mathbf{b}, \mathbf{C}\mathbf{x} \geq \mathbf{d}, \mathbf{E}\mathbf{x} = \mathbf{f}, \mathbf{x}\in\{0,1\}^l.
\end{align}
\end{subequations}
Here, $\mathbf{x}$ represents the decision variables, also known as the solution; $\mathbf{w}$ represents the problem parameters; $J(\mathbf{x}, \mathbf{w})$ is the objective function (for maximization problems, we minimize the negative objective); the elements in $\mathbf{A},\mathbf b,\mathbf C,\mathbf d,\mathbf E,\mathbf f$ are non-negative, and the three sets of constraints may not all exist simultaneously. The input to a CO problem is $\mathbf{w}$ and the constraints, and the output is a solution $\mathbf{x}$ that (as much as possible) minimizes $J(\mathbf{x}, \mathbf{w})$ within a given time.

\subsection{Framework Building Blocks}
This section describes the building blocks of the non-autoregressive CO solving framework shown in Figure~\ref{fig:overview_train}.

\textbf{Graph Modeling:} Neural networks struggle to process inputs directly in the mathematical form of (\ref{eq:formulation}), so a general solution is to model the problem parameters as a graph. Different problems have different graph modeling methods, but there is a general best practice: decision variables must correspond to node classification or edge classification tasks on the graph. For single-graph CO problems (e.g., graph cuts, node covers, TSP), the modeling process is relatively straightforward~\cite{KhalilNIPS17,kool2018attention,WangNIPS21}; if there are two graphs (e.g., graph matching), an auxiliary graph that integrates the structures of both graphs can be constructed~\cite{WangPAMI21}; for Boolean satisfiability problems, there is also a corresponding graph modeling method, where the conjunctive normal form is represented as a bipartite graph~\cite{YouNIPS19}; for general matrix-form problems (e.g., linear integer programming), a common approach is to construct a bipartite graph with $\mathbf{A}\mathbf{x} \leq \mathbf{b}$ constraints~\cite{GasseNIPS19}. In summary, for most CO problems, there exists a reasonable graph modeling method. Readers are encouraged to refer to the literature mentioned above for specific problem modeling methods.

\begin{figure}[t]
	\begin{center}
		\includegraphics[width=0.48\linewidth,page=3]{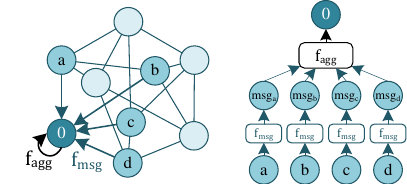}
	\end{center}
	\caption{The general framework of graph neural networks.}
	\label{fig:gnn}
\end{figure}

\textbf{Graph Neural Networks (GNNs):} Node classification or edge classification problems on graphs are naturally handled by GNNs~\cite{KipfICLR17}. There are many GNN variants, but node classification GNNs typically follow the general framework shown in Figure~\ref{fig:gnn}~\cite{YouNIPS20}: in each layer of a GNN, when node 0 needs to be updated, it only considers its neighboring nodes (a, b, c, d). For each neighbor, a message function ($f_{msg}$, typically a multi-layer neural network) transforms its current node feature into a message (e.g., $msg_a$). Next, an aggregation function ($f_{agg}$, which could be mean, sum, or attention mechanism~\cite{vaswani2017attention}) aggregates the messages from all neighbors into the current node. In every GNN layer, the weights of $f_{msg}$ and $f_{agg}$ are shared across all nodes. The choice of $f_{msg}$ and $f_{agg}$ determines the GNN variant used, with common choices including GCN~\cite{KipfICLR17}, GraphSage~\cite{HamiltonNIPS17}, and GIN~\cite{XuICLR19}. For edge classification problems, features can be propagated to edges from neighboring nodes, or node classification can be performed on the dual graph.

\textbf{Latent Code:} For problems modeled as node classification, if the decision variable dimension is $l$, a GNN with a Sigmoid activation function will output a vector of dimension $l$ with values in $[0,1]$, referred to as latent code. The $l$-dimensional real-valued space forms a latent space, where each point corresponds to a distribution of feasible solutions. This mapping is achieved using Gumbel reparameterization and the LinSAT constraint layer. Importantly, the latent space is continuous; by searching and optimizing within the latent space, better decision variables can be found.

\textbf{Gumbel Reparameterization:} The purpose of this step is to treat the latent code as a distribution over feasible solutions and sample from this distribution to obtain feasible solutions. However, discrete sampling is not differentiable, so we use Gumbel reparameterization~\cite{JiangICLR17,menaICLR18learning,GroverICLR19} to approximate discrete sampling. Given the Gumbel distribution
\begin{equation}
    g_\sigma(u) = -\sigma \log(-\log(u)),
\end{equation}
where $\sigma$ controls variance and $u$ is sampled from a uniform $(0,1)$ distribution, we apply this to the GNN's output $\mathbf{y}\in\mathbb{R}^l$:
\begin{equation}
\label{eq:gumbel}
    \widetilde{\mathbf{y}} = \left[
         \mathbf{y}_1 + g_\sigma(u)_1,
         \mathbf{y}_2 + g_\sigma(u)_2,
         \cdots,
         \mathbf{y}_l + g_\sigma(u)_l
    \right]^\top,
\end{equation}
where all $g_\sigma(u)_i$ are independent and identically distributed. To better estimate the feasible solution distribution, this sampling process needs to be repeated. Sampling can be done in parallel using batch processing on GPUs. More samples provide a more accurate estimate of the distribution but increase computational cost.

\textbf{LinSAT Constraint Layer:} After Gumbel reparameterization, we project $\widetilde{\mathbf{y}}$ into the constraint space:
\begin{equation}
    \mathbf{x} = \text{LinSAT}(\widetilde{\mathbf{y}}, \mathbf{A},\mathbf b,\mathbf C,\mathbf d,\mathbf E,\mathbf f),
\end{equation}
resulting in $\mathbf{x}\in[0,1]^l$, a quasi-discrete solution that remains continuous and differentiable. As introduced in \citet{WangICML23}, LinSAT ensures that the solution satisfies the given constraints, while Gumbel reparameterization ensures that the output is quasi-discrete without cutting off gradients. The larger the Gumbel coefficient $\sigma$, the closer the output is to a discrete solution.

The underlying mechanism of LinSAT is based on optimal transport. Optimal transport studies how to transform one marginal distribution $\mathbf{u}$ into another marginal distribution $\mathbf{v}$ with minimal cost, given a distance matrix. By constructing appropriate marginal distributions, we find that the projection process for a single linear constraint can be formalized as an optimal transport problem. For instance, for the constraint $\mathbf{a}^\top \mathbf{x} \leq b$, a matrix $\Gamma$ satisfying the following distribution is a feasible projection (with $\delta$ being dummy variables):
\begin{equation}
    \Gamma = \left[
    \begin{array}{ccccc}
         \cline{1-4}
         \multicolumn{1}{|c}{{x}_1} & {x}_2 &...& {x}_l & \multicolumn{1}{|c}{\delta_{0}}  \\
         \cline{1-4}
         \delta_{1}  & \delta_{2}  &... & \delta_{l}  & \delta_{l+1}
    \end{array}
    \right]; \quad \mathbf{u} = \underbrace{\left[a_1 \quad a_2 \quad ...\quad a_l \quad b\right]}_{l\text{-d variables}+1 \text{ dummy}}, \ 
    \mathbf{v} = \left[
    \begin{array}{c}
         b  \\
         \sum_{i=1}^l a_i 
    \end{array}
    \right],
    \label{eq:linsat}
\end{equation}
where optimal transport ensures the row sums equal $\mathbf{v}$ and the column sums equal $\mathbf{u}$. For the other two constraints $\mathbf{c}^\top \mathbf{x} \geq d$ and $\mathbf{e}^\top \mathbf{x} = f$, a similar optimal transport form can be constructed. The optimal transport problem can be solved approximately using a differentiable algorithm~\cite{SinkhornAMS64}, naturally compatible with neural networks. LinSAT extends the optimal transport algorithm to handle multiple sets of constraints simultaneously while maintaining differentiability and convergence.

\textbf{Objective Function Estimation:} One of the key advantages of non-autoregressive networks is their ability to directly estimate the objective function, which can be used as the loss function for unsupervised learning (assuming minimization problems):
\begin{equation}
    \mathcal{L}(\theta) = \mathbb{E}_{u\sim \mathcal{N}(\mathbf{0}^l, \mathbf{I}^{l\times l})}\left[ J(f(u, \mathbf{y}_\theta), \mathbf{w})\right],\label{eq:obj_est}
\end{equation}
where $\mathbf{y}_\theta$ is the GNN's output, and $f(\cdot)$ represents Gumbel reparameterization (with $u$ sampled from a standard normal distribution) and the LinSAT layer. Equation~(\ref{eq:obj_est}) computes the mean objective value under the Gumbel distribution, providing a more accurate estimate than directly using the network's output. Once $\mathbf{x} = f(u,\mathbf{y}_\theta)$ is obtained, computing $J(\mathbf{x},\mathbf{w})$ is straightforward. For linear objective functions like $J(\mathbf{x},\mathbf{w}) = \mathbf{x}^\top\mathbf{w}$, it can be directly computed. However, some CO problems involve more complex objectives, such as min/max operations, which should be avoided in practice because they truncate gradients. In the facility location problem discussed in Section~\ref{sec:exp_flp}, the objective function $\sum_{j=1}^m \min(\{\Delta_{i,j} | \ \forall \mathbf{x}_i =1\})$ can lead to poor results. Replacing $\min$ with a softmin operator, such as a softmax with temperature $\beta$, significantly improves performance. The best practice for objective function estimation is to retain gradients as much as possible and replace min/max operators with softmin/softmax to ensure smoothness.

\textbf{Unsupervised Offline Pretraining:} The differentiable nature of non-autoregressive networks supports unsupervised pretraining. During training, the mean objective function in equation~(\ref{eq:obj_est}) can serve as an unsupervised loss function. The forward propagation during training is shown with black arrows in Figure~\ref{fig:overview_train}, and backpropagation is shown with brown arrows. By minimizing this loss function during training and updating the GNN weights, the network learns to map the problem's graph representation to a good latent space.

\textbf{Online Gradient Search and Inference:} A single network output often cannot provide a sufficiently good solution, and several studies highlight the necessity of further optimization and search based on the neural network's output~\cite{SunICML23,qiu2022dimes,ChooNIPS22,joshi2019efficient,KaraliasNIPS20,abbas2022doge}. Our non-autoregressive network paradigm supports classic search methods like beam search and Monte Carlo tree search (MCTS). More importantly, the differentiable nature of our framework provides the gradient direction to optimize the objective function in the latent space, and the parallel nature allows both forward computation and gradient propagation to be done efficiently on GPUs. During the inference stage, with the network parameters $\theta$ fixed, equation~(\ref{eq:obj_est}) can be rewritten as
\begin{equation}
    \mathcal{L}(\mathbf{y}) = \mathbb{E}_{u\sim \mathcal{N}(\mathbf{0}^l, \mathbf{I}^{l\times l})}\left[ J(f(u, \mathbf{y}), \mathbf{w})\right], \label{eq:obj_search}
\end{equation}
where the gradient is propagated to update the latent code $\mathbf{y}$ rather than the network parameters. This is analogous to a new neural network training problem, where the goal is to optimize and search for better solutions via gradient optimization in the latent space. As shown in Figure~\ref{fig:overview_test}, for a specific problem instance, better latent code can be found through this process. Compared to the full dataset training, optimization on a single problem instance is more efficient and converges faster. Additionally, problem-specific neighborhood search algorithms can be designed to quickly search around the current best solution in each iteration. Experiments show that, when combined with offline training, this gradient-based search method can rival or even outperform professional solvers like SCIP and Gurobi.

\subsection{Further Discussions}
\subsubsection{Expressive Power of Graph Neural Networks for Combinatorial Optimization}
GNNs were initially developed to process "common" graph structures such as social networks. However, current theoretical studies on GNN expressivity often begin with graph isomorphism problems, which are a form of CO. The mainstream view is that the expressivity of GNNs is comparable to the Weisfeiler-Lehman (WL) graph isomorphism test~\cite{WeisfeilerLemanTest1968,XuICLR19}. Although the graph isomorphism problem is also a CO problem, it should be noted that the WL test is an approximate algorithm, a sufficient but not necessary condition for graph isomorphism. Whether GNNs have the expressivity to solve any CO problem remains an open question. However, practical experience in machine learning for CO shows that GNNs are up to the task~\cite{KhalilNIPS17,GasseNIPS19,KaraliasNIPS20}. Furthermore, recent theoretical research has shown positive results. For example, Chen et al.~\cite{ChenICLR2023LP,ChenICLR2023MIP} proved that GNNs have enough expressivity to predict feasibility, optimal objective values, and optimal solutions for linear and mixed-integer programming problems. Hence, this paper chooses GNNs as the primary neural network architecture, though the framework is flexible enough to support more expressive neural networks.

\subsubsection{Advantages of Gumbel Reparameterization}
\label{sec:gumbel_strengths}
This paper adopts Gumbel reparameterization~\cite{JiangICLR17} to enable differentiable sampling in (nearly) discrete spaces, offering two main advantages.

First, \emph{Gumbel reparameterization allows more accurate gradient computation}. A defining characteristic of CO problems is that the feasible domain of decision variables is non-continuous, which conflicts with the continuous nature of neural networks. The non-continuous feasible domain presents a challenge for end-to-end gradient computation: if the network output is discrete, the input-output mapping becomes a piecewise function~\cite{PoganvcicICLR19}. A piecewise function has gradients that are impulse (delta) functions, providing little meaningful information for the network. Discarding the discrete part of the constraints retains gradients, but this turns the training problem into a relaxed continuous optimization problem. In many cases, this is a simpler convex optimization problem, fundamentally different from the actual CO problem, and a network trained on convex optimization may not generalize to more difficult CO problems. By introducing Gumbel reparameterization, we ensure that the network outputs are as close to the discrete domain as possible without cutting off gradients, resulting in more accurate gradients during optimization.

Second, \emph{Gumbel reparameterization enables parallel sampling during inference}. In this paper, each latent variable corresponds to a distribution over feasible solutions. During inference, Gumbel reparameterization samples from this distribution. More precise gradients lead to faster latent variable updates during optimization, and the randomness introduced by Gumbel reparameterization gives the network a higher chance of finding better solutions. Furthermore, the sampling process is parallelizable, and batch operations allow it to be efficiently executed on GPUs, making it more efficient than traditional sampling and search algorithms~\cite{WangPAMI21,ningxiuPhDThesis}.

\subsubsection{Revisiting Deep Learning Frameworks}
The success of deep learning in recent years is largely due to the development of GPU computing and efficient open-source deep learning frameworks (especially the highly optimized low-level implementations). One reason for the success of the proposed non-autoregressive framework is its ability to leverage these high-efficiency deep learning frameworks. With good support for GPUs and efficient low-level code, operations like Gumbel sampling and objective function estimation for multiple quasi-discrete solutions can be performed in parallel on GPUs. Additionally, deep learning frameworks provide highly efficient large-scale gradient optimization algorithms. Our framework reuses the automatic differentiation capabilities of deep learning frameworks, especially during inference, where automatic differentiation is used to update latent code, enabling efficient online gradient-based search.

\subsubsection{A Meta-Learning Perspective on Combinatorial Optimization}
Qiu et al.~\cite{qiu2022dimes} were the first to propose a meta-learning perspective (more precisely, model-agnostic meta-learning~\cite{FinnICML17}) for CO. In the pretraining stage, the neural network learns a set of initial weights that are effective across the entire dataset. In specific CO problems, the network weights can be further updated to obtain better solutions. Our framework can also be understood from a similar meta-learning perspective, but here the meta-learning object shifts from the neural network weights to the lighter latent code: in the pretraining stage, the network outputs learn to map to a set of latent codes that are effective across the entire dataset; for specific CO problems, the latent code can be further optimized via gradient descent to obtain better solutions.

\section{Experiments and Analysis}
\label{sec:experiment}
This paper demonstrates the effectiveness of non-autoregressive networks on facility location, max-set covering, and traveling salesman problems, comparing our approach to existing neural network methods~\cite{KaraliasNIPS20,WangICLR23} and traditional MILP solvers~\cite{SCIP,gurobi}. In the facility location and max-set covering experiments, the primary goal is to compare the performance of non-autoregressive neural networks with traditional solvers (integer programming solvers like SCIP and Gurobi); in the traveling salesman problem, the focus is on comparing our framework with other neural network-based solvers. Facility location and max-set covering experiments were run on a workstation with an i7-9700K CPU, 16GB RAM, and an RTX 2080Ti GPU, while the traveling salesman problem experiments were conducted on a workstation with an AMD 3970X CPU, 32GB RAM, and an RTX 3090 GPU.

\subsection{Facility Location Problem}
\label{sec:exp_flp}
The facility location problem (FLP) is formulated as follows: Given $m$ locations, we are to choose $k$ locations to build new facilities, where each facility will serve other locations based on proximity (subject to capacity limits). The objective is to minimize the sum of distances from each location to the nearest facility. In the simplified version without capacity limits, the problem is formulated as:
\begin{equation}
\label{eq:flp}
    \min_{\mathbf{x}} \ \sum_{j=1}^m \min(\{{\Delta}_{i,j} | \ \forall \mathbf{x}_i =1\}), \quad
    s.t. \quad \mathbf{x}\in\{0,1\}^{m}, \ \sum_{i=1}^m\mathbf{x}_i \leq k,
\end{equation}
where $\mathbf{x}$ represents the decision variables, and $\Delta_{i,j}$ means the distance between locations $i$ and $j$. For this problem, the implementation of the non-autoregressive neural network is as follows:

\begin{figure}[tb!]
\centering
\captionsetup[subfigure]{aboveskip=-5pt,belowskip=1pt}
\begin{subfigure}[b]{0.38\textwidth}
    \includegraphics[width=\textwidth]{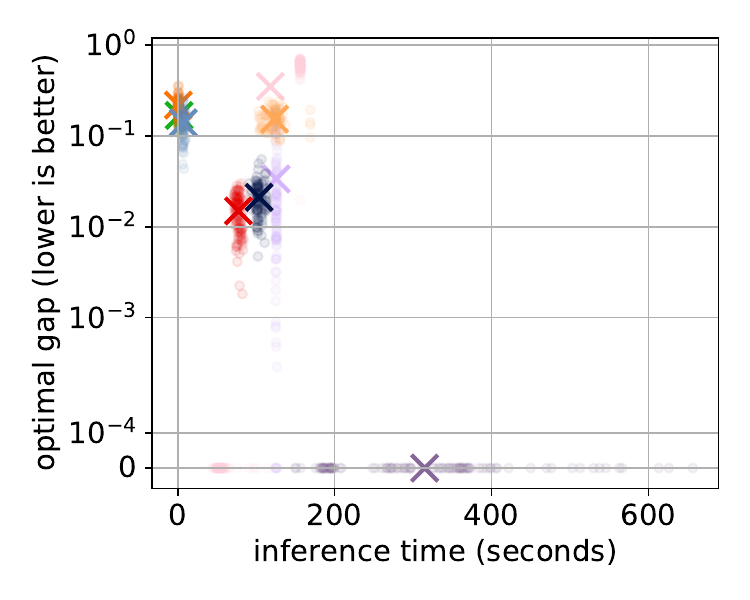}
    \caption{FLP (k=30, m=500)}
\end{subfigure}
\begin{subfigure}[b]{0.38\textwidth}
    \includegraphics[width=\textwidth]{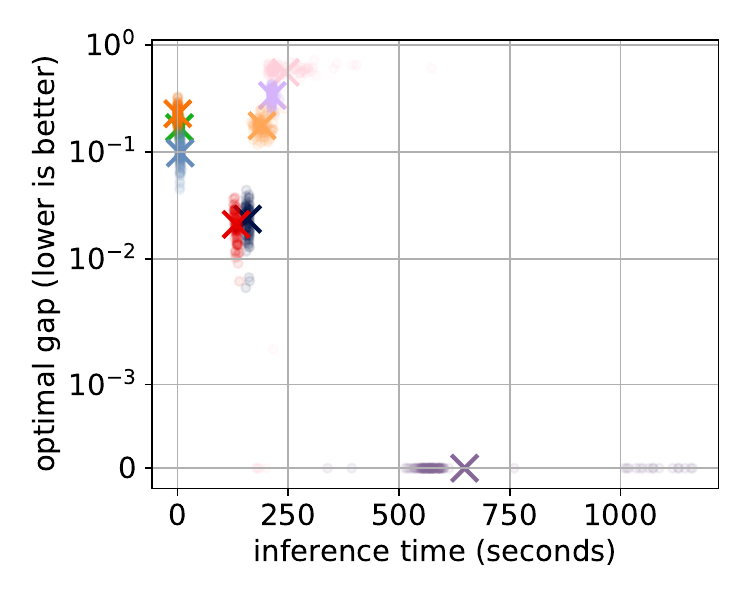}
    \caption{FLP (k=50, m=800)}
\end{subfigure}
    \includegraphics[width=0.19\textwidth]{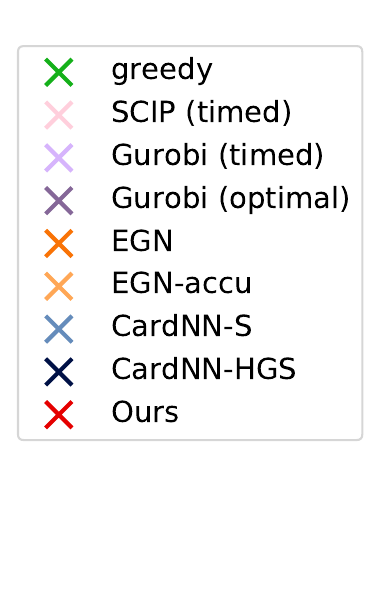}
\caption{Comparison on gap w.r.t.\ optimal solution on uncapacitated facility location problem.}
\label{fig:flp}
\end{figure}

\noindent $\quad\bullet$ \emph{Graph Modeling:} The problem is modeled as a graph, where each location is a node, and edges are defined between locations with a distance less than 2\% of the total area diameter. The facility location problem is equivalent to node classification on this graph.

\noindent $\quad\bullet$ \emph{Graph Neural Network:} A spline convolutional network~\cite{FeyCVPR18} is used, with 3 layers and a hidden dimension of 16 and 5 spline kernels.

\noindent $\quad\bullet$ \emph{Gumbel Reparameterization and LinSAT Layer:} These steps ensure that the network output is projected into the feasible region with constraints from (\ref{eq:flp}).

\noindent $\quad\bullet$ \emph{Objective Function Estimation:} Since the $\min$ operator in (\ref{eq:flp}) truncates gradients, we use a smooth softmin operator (i.e., applying softmax after negating the input). Specifically, we replace $\min(\{\Delta_{i,j}|\ \forall \mathbf{x}_i=1\})$ with $\text{softmax}(-\beta {\Delta} \circ \mathbf{x})$, where $\circ$ denotes element-wise multiplication, and $\beta$ is a temperature parameter.

\begin{figure}[tb]
\centering
\centering
\includegraphics[width=0.38\textwidth]{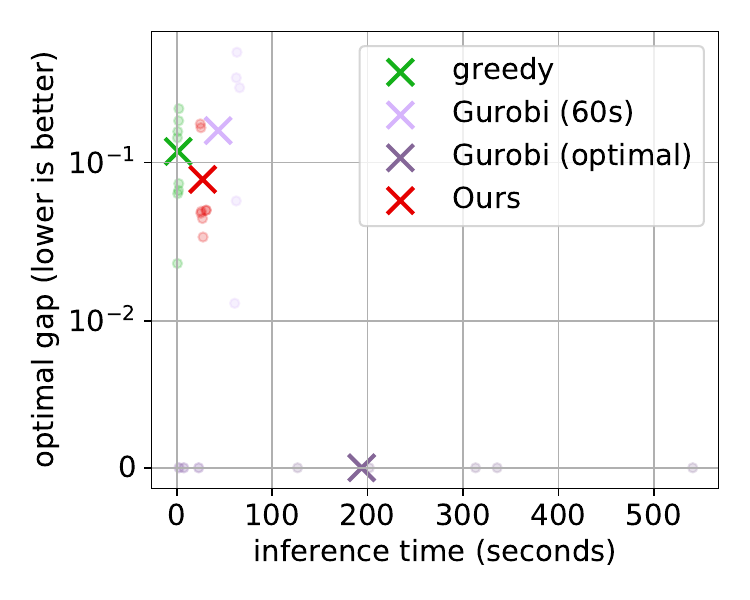}
\caption{Generalization study in real-world.}
\label{fig:flp_realworld}
\end{figure}

\noindent $\quad\bullet$ \emph{Neighborhood Search in Inference:} In the facility location problem, we apply k-median clustering for fast search around the best solutions found in the batch.

We follow the experiment design in~\cite{WangICLR23}, where $m$ locations are randomly generated in a 2D unit area to form training/testing data. All neural methods were trained without any optimal solution information during training. In the experiments, we compare our method with greedy algorithms and professional solvers (SCIP~\cite{SCIP} and Gurobi~\cite{gurobi}), with time limits of 120 seconds and 200 seconds for Figures~\ref{fig:flp}(a) and (b), respectively. We also compare with neural solvers such as EGN~\cite{KaraliasNIPS20} and CardNN~\cite{WangICML23}, with results shown in Figure~\ref{fig:flp}. The figure compares different solvers in terms of both solution quality (measured by the gap relative to the optimal solution) and runtime, with each circle representing a test instance, and the average performance of each solver shown by an "X". Our non-autoregressive neural solver (red points) achieves Pareto-optimality in terms of both solution accuracy and runtime, surpassing even commercial solvers like Gurobi in both efficiency and performance.

To test the generalization ability of the neural solver to unseen distributions, we constructed a real-world dataset: using the actual geographic coordinates of all Starbucks stores in Shanghai, New York, London, and Seoul as inputs, calculating Euclidean and Manhattan distances, and solving the facility location problem. The neural network was trained on simulated data but tested directly on real data. As shown in Figure~\ref{fig:flp_realworld}, the neural network exhibits strong generalization, outperforming Gurobi on unseen real-world data. Tables~\ref{tab:flp_ablation1} and~\ref{tab:flp_ablation2} provide ablation studies on the main components of our framework, demonstrating the importance of online gradient search and inference, as well as the significance of softmin operators and hyperparameter selection for the facility location problem.

\textbf{Capacitated Facility Location Problem.} To further demonstrate the practicality of our method, we also consider the capacitated facility location problem, where each facility can only serve a limited number of locations. The optimization problem is formulated as:
\begin{subequations}
\label{eq:flp_cap}
\begin{align}
    \min_{\mathbf{x}} \ \min_{\mathbf{P}} \sum_{i=1}^m\sum_{j=1}^m \Delta_{i,j} \mathbf{P}_{i,j},
    \quad s.t. \quad &\mathbf{x}\in\{0,1\}^m, \sum_{i=1}^m \mathbf{x}_i \leq k,  \sum_{i=1}^m \mathbf{x}_i \geq \sum_{i=1}^m d_i, \\
    & \mathbf{P} \in \mathbb{R}_+^{m\times m}, \sum_{i=1}^m\mathbf{P}_{i,j} = d_j, \sum_{j=1}^m \mathbf{P}_{i,j} \leq \mathbf{x}_i.  \label{eq:flp_cap_constr}
\end{align}
\end{subequations}
Here, $\mathbf{P}_{i,j}$ represents how much demand from location $j$ is served by facility $i$. The implementation for the capacitated version of the problem is largely similar to the uncapacitated case, with a few modifications:

\begin{table}[!tb]
\parbox{.53\linewidth}{
\caption{Optimal gap experiment result with different $\beta$ values on uncapacitated facility location problem (k=30,m=500).}
\label{tab:flp_ablation1}
\footnotesize
\tabcolsep 10pt 
    \centering
    \begin{tabular}{ccccc}
        \toprule
       $\beta=10$ & $\beta=20$ & $\beta=50$ & $\beta=100$ & $\beta=200$ \\
       \midrule
       0.0711 &  0.0501 & \textbf{0.0149} & 0.0428 & 0.0606 \\
       \bottomrule
    \end{tabular}
}
\parbox{.45\linewidth}{
\caption{Ablation study with fixed $\beta$.}
\label{tab:flp_ablation2}
\footnotesize
\tabcolsep 10pt 
    \centering
    \begin{tabular}{ccc}
    \cmidrule[\heavyrulewidth]{1-3} 
       Gradient search & min / softmin & Gap \\
       \cmidrule{1-3} 
       $\checkmark$  & min & 0.1772 \\
       $\times$  & softmin ($\beta=50$) & 0.1753 \\
       $\checkmark$  & softmin ($\beta=50$) & \textbf{0.0149} \\
    \cmidrule[\heavyrulewidth]{1-3} 
    \end{tabular}
}
\end{table}

\begin{table}[tb]
\caption{Results on capacitated facility location problem.}
\label{tab:flp_cap}
\footnotesize
\tabcolsep 4pt 
\centering
\begin{tabular}{ccc}
\toprule
  Objective score ($\downarrow$) & Gurobi & Ours \\\hline
  Time limit = 30s & 2.6454$\pm$0.1018 & \textbf{2.5026$\pm$0.0824}
 \\
  Time limit = 100s & \textbf{2.4524$\pm$0.0756} & 2.4797$\pm$0.0715
 \\
\bottomrule
\end{tabular}
\end{table}

\noindent $\quad\bullet$ \emph{LinSAT Layer:} The constraints for decision variables $\mathbf{x}$ are updated to account for the capacity limits, but these still fall under the class of positive linear constraints and can be handled by the LinSAT layer.\footnote{Although this problem involves two sets of decision variables ($\mathbf{x}$ and $\mathbf{P}$), the neural network only needs to predict the binary decision variables $\mathbf{x}$, as $\mathbf{P}$ can be determined by a polynomial-time algorithm (in this case, an optimal transport algorithm~\cite{cuturiNIPS13}).}

\noindent $\quad\bullet$ \emph{Objective Function Estimation:} The objective is more complex due to the capacity limits, as each location cannot simply choose the nearest facility. After determining $\mathbf{x}$, the remaining part of the problem is a linear programming problem, specifically an optimal transport problem. While linear programming problems are theoretically differentiable~\cite{AmosICML17}, solving multiple linear programs in a batch during training would be computationally expensive. We use a GPU-parallelizable approximation method~\cite{cuturiNIPS13} to solve the optimal transport problem and employ a more efficient backpropagation technique~\cite{GouldPAMI21}.

The experiments use the same data distribution as in Figure~\ref{fig:flp}(a), with each facility able to serve up to 50 locations. Table~\ref{tab:flp_cap} compares our method with Gurobi. For shorter time limits (around 30 seconds), our non-autoregressive network outperforms Gurobi; for longer time limits (around 100 seconds), Gurobi outperforms our method, though the difference is small. In this problem setting, existing neural methods like CardNN~\cite{WangICLR23} do not apply due to the constraint changes; methods like EGN~\cite{KaraliasNIPS20} and SCIP~\cite{SCIP} were omitted from Table~\ref{tab:flp_cap} for brevity, as they did not outperform the best neural network and traditional solver.

In summary, for the facility location problem, our non-autoregressive neural network solver outperforms existing (non-autoregressive) neural solvers in terms of applicability and accuracy. Its performance is comparable to, and in some cases surpasses, professional solvers like Gurobi.

\subsection{Max-Set Covering Problem}

\begin{figure}[tb!]
\centering
\captionsetup[subfigure]{aboveskip=-5pt,belowskip=1pt}
\begin{subfigure}[b]{0.38\textwidth}
    \includegraphics[width=\textwidth]{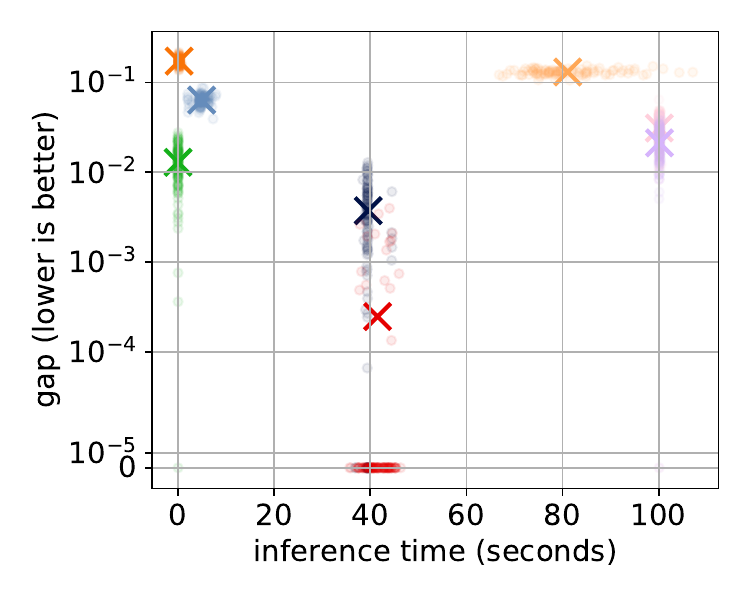}
    \caption{MCP (k=50, m=500, n=1000)}
\end{subfigure}
\begin{subfigure}[b]{0.38\textwidth}
    \includegraphics[width=\textwidth]{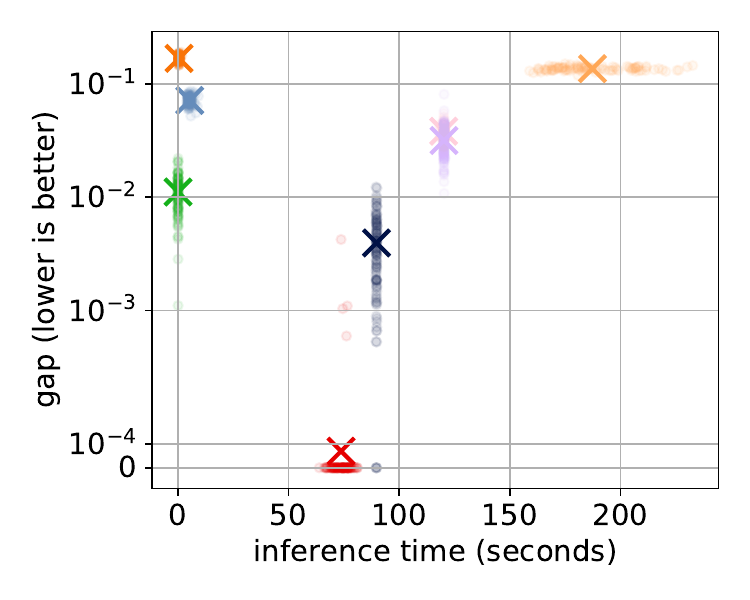}
    \caption{MCP (k=100, m=1000, n=2000)}
\end{subfigure}
    \includegraphics[width=0.19\textwidth]{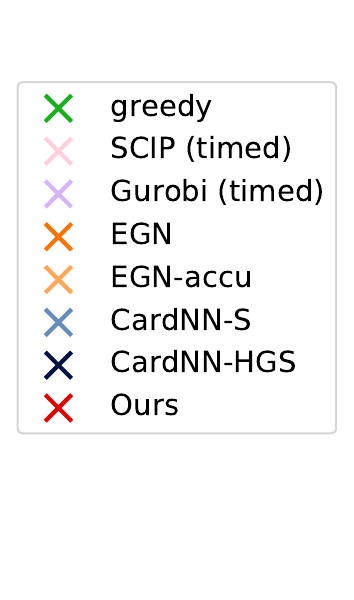}
\caption{Comparison on gap w.r.t.\ best known incumbent solution on max-set covering problem.}
\label{fig:mcp}
\end{figure}

The max-set covering problem (MCP) is formulated as follows: Given $m$ sets and $n$ items, each set covers some items from the universe. We are to select $k$ sets such that the total value of the items covered by the selected sets is maximized. A typical case of MCP arises in social networks, where sets represent $m$ influencers, and items represent $n$ users. An advertiser can solve MCP to maximize the impact of an advertisement within a limited budget. The MCP is formulated as:
\begin{align}
    \max_{\mathbf{x}} \ & \sum_{j=1}^n \left( \mathbb{I}\left(\sum_{i=1}^m \mathbf{x}_{i} \mathbf{A}_{ij}\right) \cdot \mathbf{v}_j \right), \quad
    s.t. \quad \mathbf{x} \in \{0,1\}^m, \sum_{i=1}^m\mathbf{x}_i \leq k.\label{eq:mcp}
\end{align}
Here, $\mathbf{v}\in\mathbb{R}_+^{n}$ represents the value of each item, $\mathbf{A}_{i,j}=1$ means set $i$ covers item $j$, and $\mathbb{I}(\cdot)$ is an indicator function that outputs 1 if the internal sum is greater than 1, and 0 otherwise. The implementation of our non-autoregressive network for MCP is as follows:

\noindent $\quad\bullet$ \emph{Graph Modeling:} The MCP is modeled as a bipartite graph where sets and items are the two partitions. An edge exists between set $i$ and item $j$ if $\mathbf{A}_{i,j}=1$, and solving MCP is equivalent to node classification on the "sets" partition.

\noindent $\quad\bullet$ \emph{Graph Neural Network:} A GraphSage~\cite{HamiltonNIPS17} is used, with three layers. The message passing from "sets" to "items" and vice versa is handled by two separate networks. The hidden dimension is 16.

\noindent $\quad\bullet$ \emph{Gumbel Reparameterization and LinSAT Constraint Layer:} These steps ensure that the network output satisfies the constraints in (\ref{eq:mcp}).

\noindent $\quad\bullet$ \emph{Objective Function Estimation:} Compared to the facility location problem, the objective in (\ref{eq:mcp}) is more straightforward: after predicting $\mathbf{x}$, the objective function is directly calculated based on (\ref{eq:mcp}).

\noindent $\quad\bullet$ \emph{Neighborhood Search in Inference:} MCP did not employ a specific neighborhood search in this case, but online gradient search is still utilized.

We generate two sets of training/testing data with distributions similar to ORLIB~\cite{orlib}, following the setup in~\cite{WangICLR23}. We compare our method with greedy algorithms, professional solvers (SCIP~\cite{SCIP}, Gurobi~\cite{gurobi}, with 100-second and 120-second time limits for Figures~\ref{fig:mcp}(a) and (b), respectively), and neural solvers such as EGN~\cite{KaraliasNIPS20} and CardNN~\cite{WangICML23}. The results are shown in Figure~\ref{fig:mcp}, where the gap relative to the best-known incumbent solution is compared against runtime. Each circle represents a test instance, and the average performance of each solver is denoted by an "X." In MCP, even state-of-the-art solvers like Gurobi cannot return the optimal solution within 24 hours for some instances, so the gap is measured relative to the best-known incumbent solution. Our non-autoregressive solver (purple) achieves Pareto-optimal performance, outperforming Gurobi in both runtime and solution quality. Additionally, our method finds the best-known solutions on many test instances.

\subsection{Traveling Salesman Problem}

The traveling salesman problem (TSP) is formulated as follows: Given $m$ cities with known coordinates, a traveling salesman needs to visit all cities and return to the starting city, minimizing the total travel distance. The mathematical formulation is:
\begin{equation}
    \min_\mathbf{X} \ \frac{1}{2}\mathrm{tr}(\mathbf{D}^\top \mathbf{X}), \quad s.t. \quad \mathbf{X} \in \{0,1\}^{m\times m}, \forall j : \sum_{i=1}^m \mathbf{X}_{i,j} = 2, \forall i : \sum_{j=1}^m \mathbf{X}_{i,j} = 2, \mathbf{X} \in \mathcal{H},
\end{equation}
where $\mathbf{D}\in\mathbb{R}_+^{m\times m}$ represents the pairwise distances between cities, $\mathbf{X}_{i,j}=1$ indicates that the salesman travels between cities $i$ and $j$, and $\mathbf{X} \in \mathcal{H}$ represents a Hamiltonian cycle. Our non-autoregressive solver for TSP is implemented as follows:

\noindent $\quad\bullet$ \emph{Graph Modeling:} The cities' coordinates are used as node features, and pairwise distances are used as edge features to construct a fully connected graph. For the larger TSP-500 task, the graph is sparsified using a $k$-nearest neighbors approach ($k=50$).

\noindent $\quad\bullet$ \emph{Graph Neural Network:} We use an anisotropic graph neural network~\cite{WangICDE12}, which, unlike traditional GNNs~\cite{YouNIPS20}, encodes both node and edge features and performs message passing between nodes and edges.

\noindent $\quad\bullet$ \emph{LinSAT Layer:} The Hamiltonian cycle constraint cannot be directly handled by LinSAT, but its relaxed form $\forall j: \sum_{i=1}^m \mathbf{X}_{i,j} = 2, \forall i: \sum_{j=1}^m \mathbf{X}_{i,j} = 2$ is a positive linear constraint that can be handled by LinSAT. Gumbel reparameterization is not used in this experiment due to conflicts with the existing TSP framework.

\noindent $\quad\bullet$ \emph{Supervised Network Training:} To ensure a fair comparison with other neural solvers for TSP, we follow supervised training practices from the literature~\cite{sun2023difusco,li2023t2t}, unlike the unsupervised learning used in other experiments in this paper.

\noindent $\quad\bullet$ \emph{Neighborhood Search in Inference:} For TSP, we apply greedy output combined with 2-Opt search and Monte Carlo tree search (MCTS) strategies, which are commonly used in solving TSP.

\begin{table}[t]
    \centering
    \caption{Our neural solver performs on-par or better than other neural networks for TSP.}
        \label{tab:tsp}
    \resizebox{\textwidth}{!}{
    \begin{tabular}{clccccccccc}
        \toprule
        \multicolumn{2}{c}{Solver Type} & \multicolumn{3}{c}{TSP-50} & \multicolumn{3}{c}{TSP-100} & \multicolumn{3}{c}{TSP-500} \\
        \cmidrule(lr){1-2} \cmidrule(lr){3-5} \cmidrule(lr){6-8} \cmidrule(lr){9-11}
        Neural Network & Search Strategy & Obj ($\downarrow$) & Gap ($\downarrow$) & Time ($\downarrow$) & Obj ($\downarrow$) & Gap ($\downarrow$) & Time ($\downarrow$) & Obj ($\downarrow$) & Gap ($\downarrow$) & Time ($\downarrow$) \\
        \midrule
        \multicolumn{2}{c}{LKH3 (max\_trials = 500 | 5k | 50k)} & 5.688 & 0.00\% & 3m12s & 7.756 & 0.00\% & 33m18s & 16.548 & 0.00\% & 56m12s \\
        DIMES~\cite{qiu2022dimes} & Active Search+Sampling & 5.735 & 0.82\% & 1h44m & 7.918 & 2.09\% & 3h19m &  17.646 &  6.64\% & 9h42m\\
        DIFUSCO~\cite{sun2023difusco} & Greedy+2Opt & 5.694 & 0.11\% & 8m24s & 7.776 & 0.26\% & 8m44s & 16.800 & 1.52\% & 2m34s  \\
        T2T~\cite{li2023t2t} & Greedy+2Opt & 5.689 & 0.02\% & 25m12s & 7.761 &  0.07\% & 25m58s & 16.695 & 0.89\% & 6m3s \\
        \midrule
        GNN & Greedy+2Opt & 5.694 & 0.12\% & 15s & 7.806 & 0.65\% & 20s & 16.899 & 2.12\% & 15s \\
        GNN+LinSAT (Ours) & Greedy+2Opt & 5.691 & 0.07\% & 18s & 7.777 & 0.27\% & 25s & 16.746 & 1.19\% & 18s \\
        GNN & Greedy+MCTS & 5.689 & 0.02\% & 15s & 7.767 & 0.01\% & 29s & 16.706 & 0.95\% & 1m4s \\
        GNN+LinSAT (Ours) & Greedy+MCTS & 5.689 & 0.03\% & 27s & 7.771 & 0.20\% & 2m8s & 16.646 & 0.59\% & 1m8s \\
        \bottomrule
    \end{tabular}
    }
\end{table}

As shown in Table~\ref{tab:tsp}, we validate the effectiveness of our method on TSP instances with $m=50, 100, 500$. Following common practice in the field, the city coordinates are randomly generated in a 2D uniform distribution in $[0,1]$, and the Euclidean distance between each pair of cities is computed. The TSP-50 and TSP-100 datasets each have 1280 test instances, while TSP-500 has 128 test instances. The reported times in Table~\ref{tab:tsp} are the total runtime across the entire test set. In terms of both solution quality and runtime, our non-autoregressive method performs on par or better than existing neural network solvers~\cite{qiu2022dimes,sun2023difusco,li2023t2t}. TSP is one of the most well-researched CO problems in machine learning~\cite{VinyalsNIPS15,KhalilNIPS17}, and the primary purpose of this experiment is to compare our neural network (particularly the LinSAT-augmented non-autoregressive network) against other state-of-the-art neural solvers. In Table~\ref{tab:tsp}, baseline methods like~\cite{qiu2022dimes,sun2023difusco,li2023t2t} employ more complex deep learning architectures, such as meta-reinforcement learning and diffusion models. We also compare the performance of our method with and without LinSAT. Without LinSAT, the neural network performs better on smaller TSP instances but struggles with larger instances. This indicates that explicitly considering constraints (even in their relaxed form) improves the model's capacity, as without this, additional parameters would be needed to learn the constraints. However, it should be noted that traditional solvers like LKH3 still outperform neural network methods in terms of solution quality. This is not surprising given that LKH has been fine-tuned for decades for the TSP problem specifically. However, in many other CO problems, the TSP is an outlier, and Gurobi is typically the default general solver. Based on this, we use LKH3 as the benchmark in Table~\ref{tab:tsp}.

\section{Conclusion and Outlook}
The success of neural networks has been complemented by advancements in deep learning frameworks and GPU hardware. Given the fundamental importance of CO, leveraging the computational power of neural networks (especially deep learning frameworks and GPUs) for CO is becoming a cutting-edge research direction. This paper first summarizes existing autoregressive neural solvers, which can better handle constraints and train using reinforcement learning. However, autoregressive networks face challenges like error accumulation, sparse rewards, reduced efficiency, and the inability to model permutation invariance in large-scale problems. The development of techniques like differentiable constraint encoding, Gumbel reparameterization, and unsupervised learning makes non-autoregressive networks a promising research direction.

In pursuit of a general-purpose non-autoregressive neural solver for CO, this paper proposes a framework, including building blocks such as graph modeling, GNNs, latent code (latent variables), Gumbel reparameterization, LinSAT layers, and (differentiable) objective function estimation. Our approach can handle all CO problems with positive linear constraints, which is more general than other non-autoregressive methods. Our framework supports offline unsupervised pretraining, reducing the need for labeled data and making it more practical. We also present an online gradient search technique that enhances the network's generalization ability. Our method was validated on (capacitated/un-capacitated) facility location, max-set covering, and TSP problems, surpassing existing non-autoregressive networks in terms of applicability and performance. For TSP, our LinSAT-augmented non-autoregressive network performed on par with other state-of-the-art neural methods; for facility location and max-set covering, our method achieved comparable performance to commercial solvers like Gurobi and even outperformed them on certain problem instances.

We believe that the advantage of neural solvers (especially non-autoregressive networks) lies in offering optimization solvers whereby the user has full control of. All building blocks introduced in this paper are open-sourced. In real-world applications, data distributions are often specific and limited, so sacrificing some generality for better performance within a particular distribution is usually acceptable. Finally, this paper's experiments have not fully explored the design space of neural networks, customized search algorithms, or just-in-time Python compilation techniques, and we believe there is room for further improvement in the performance of neural solvers. The research into non-autoregressive networks for CO optimization is still in its infancy, with many opportunities to improve generality, efficiency, and explore more applications.

{
\bibliographystyle{gbt7714-numerical}
\bibliography{egbib}
}

\newcommand*{\authorcv}[3][]{
    \def\@authorphoto{#1}
    \noindent
    \begin{minipage}[t]{0.45\textwidth}
        \renewcommand{\baselinestretch}{1.25}
        \baselineskip 9pt
        \parindent=9pt
        \ifx\@authorphoto\@empty
        \else \parpic{\includegraphics[width=25mm]{#1}}
        \fi
        {\noindent{\bf #2}~#3\par}
    \end{minipage}
}

\hspace*{-4em}
\begin{tabular*}{\textwidth}{cc}
\authorcv[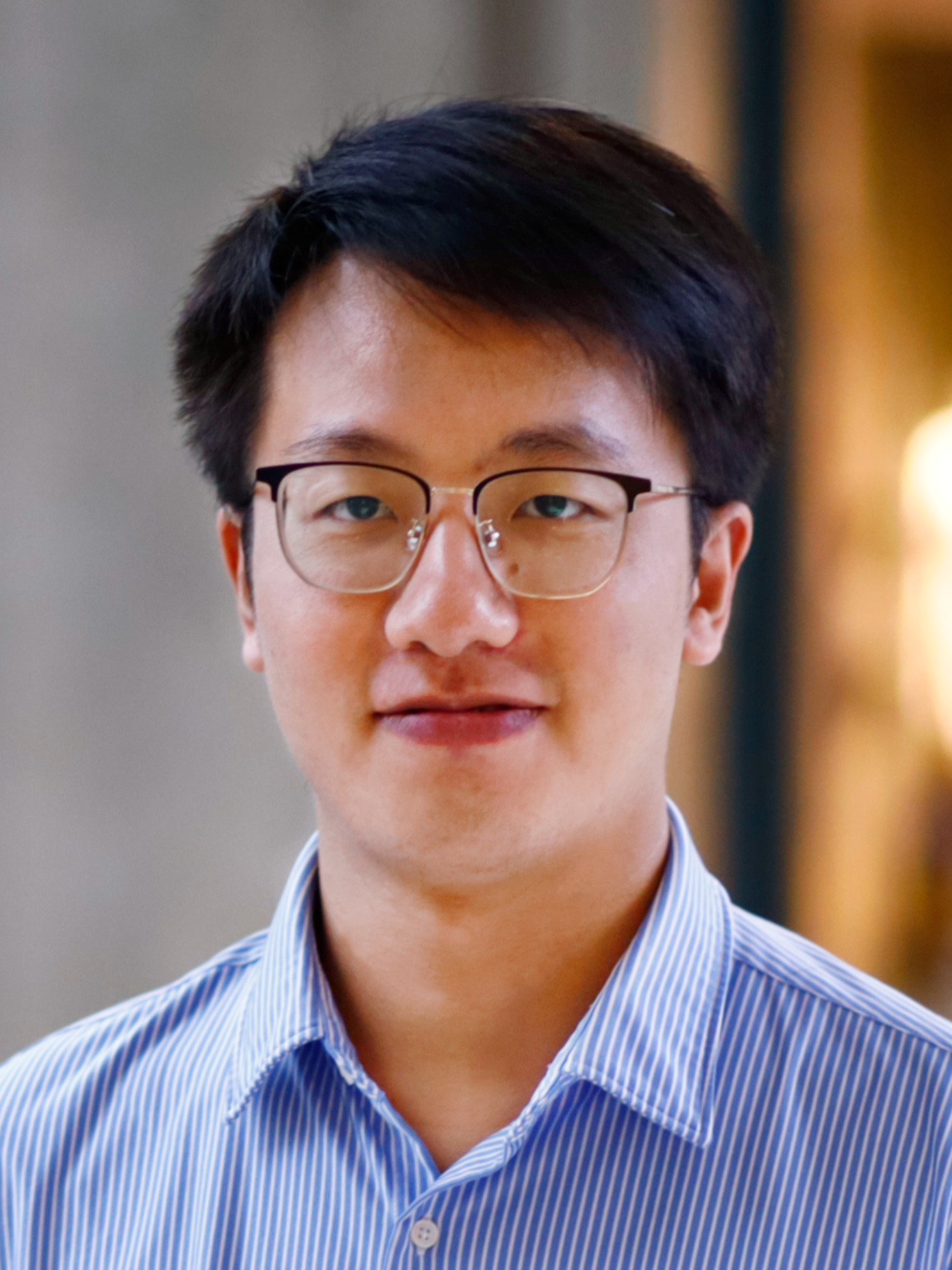]{Runzhong WANG}{received his PhD degree in Computer Science, Shanghai Jiao Tong University in 2023. He obtained B.E. in Electrical Engineering from Shanghai Jiao Tong University in 2019. He has published 10+ first-authored papers in NeurIPS, ICML, ICLR, IEEE TPAMI, etc.\ on learning for combinatorial optimization. He is maintaining several learning combinatorial optimization repositories with 2000+ stars and currently is a postdoc associate at MIT, USA.} &

\authorcv[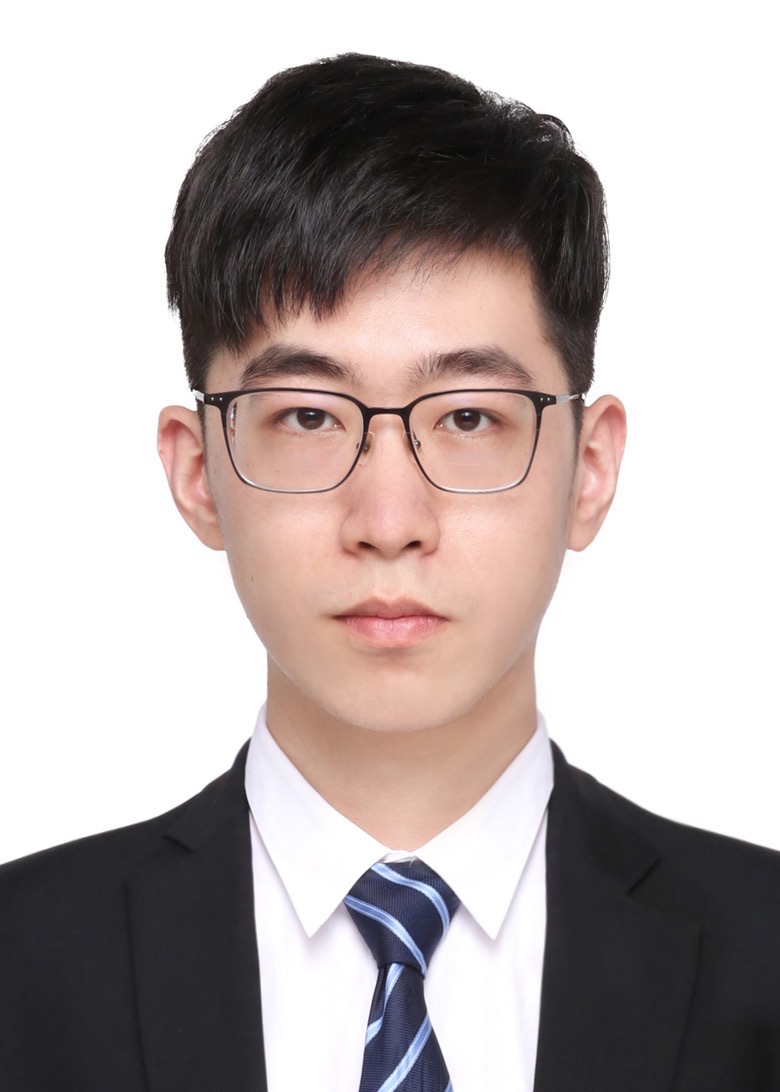]{Yang LI}{is a second-year master student in Computer Science, Shanghai Jiao Tong University. He obtained B.E. in Cyber Science and Engineering from Shanghai Jiao Tong University in 2022. His research interests lie in learning for combinatorial optimization and generative models. He has published 5 first-authored papers in NeurIPS, ICLR, KDD, etc.\ and serves as a reviewer for NeurIPS, ICLR, ICML, etc.} \\

\authorcv[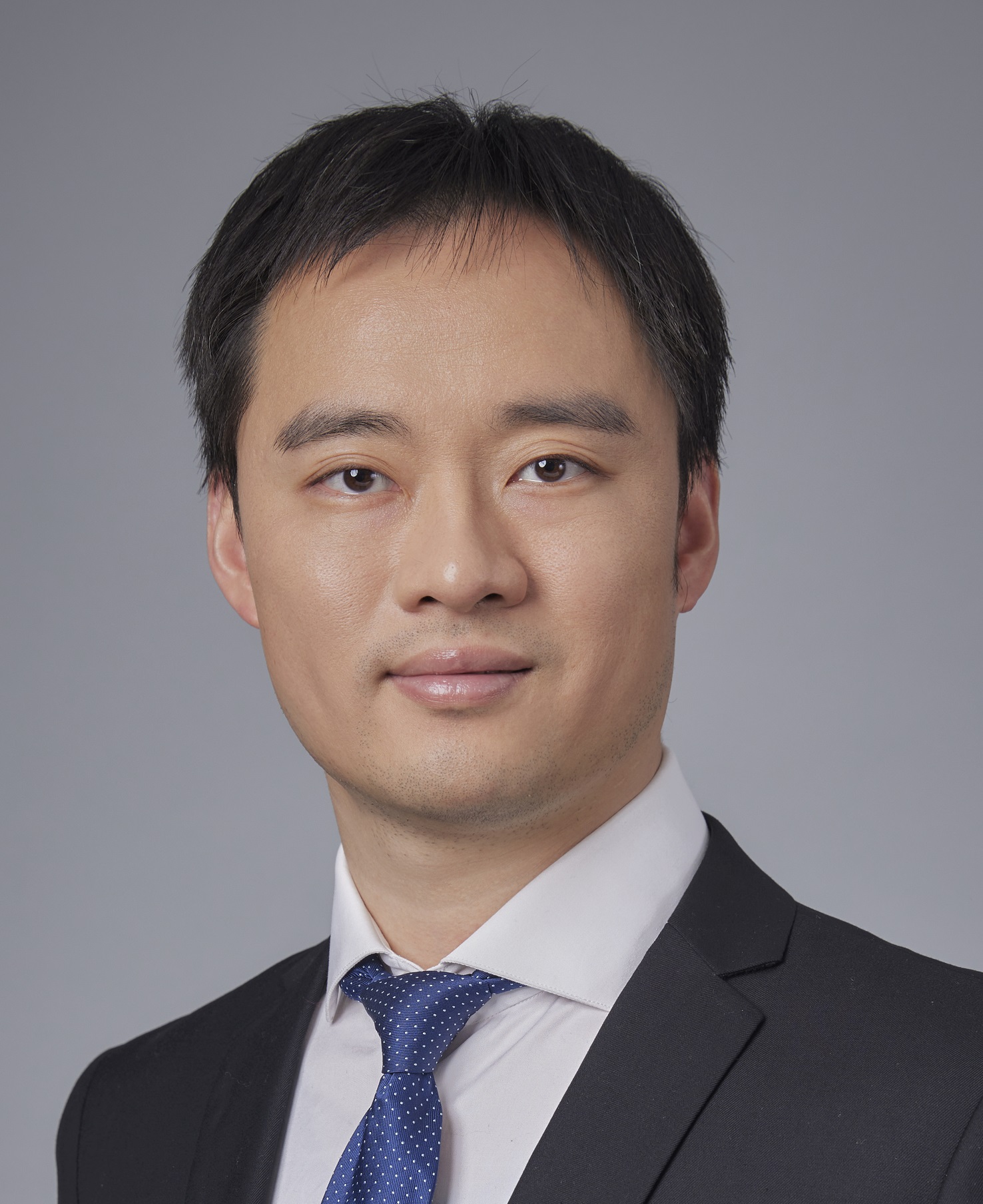]{Junchi YAN}{is a Professor with School of Artificial Intelligence and Department of Computer Science and Engineering, Shanghai Jiao Tong University. He was a Research Staff Member with IBM. His research interests include machine learning and applications. He served Area Chair for ICML/ICLR/NeurIPS etc., and Associate Editor for IEEE TPAMI, Pattern Recognition. He is a Fellow of IET, and a Senior Member of IEEE.} &

\authorcv[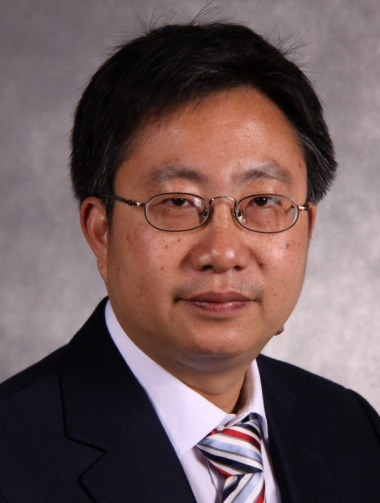]{Xiaokang YANG}{received the B. S. degree from Xiamen University, in 1994, the M. S. from Chinese Academy of Sciences in 1997, and the Ph.D. from Shanghai Jiao Tong University in 2000. He is currently a Distinguished Professor and the Executive Director of AI Institute, Shanghai Jiao Tong University, Shanghai, China. His research interests are mainly image processing and computer vision. He is a Fellow of IEEE.}
\end{tabular*}

\end{document}